\documentclass[sigconf]{acmart}
\pdfoutput=1

\usepackage{booktabs} 
\usepackage{bm}
\usepackage{subcaption}
\usepackage{mathrsfs}
\usepackage[ruled]{algorithm2e}
\usepackage{amsmath}
\usepackage{caption}
\usepackage{enumitem}
\usepackage{comment}
\usepackage{graphicx}
\usepackage{amsthm}
\usepackage{threeparttable}
\usepackage{multirow}
\usepackage{color}

\newtheorem{theorem}{Theorem}[section]

\newtheorem{definition}[theorem]{Definition}

\newtheorem*{conjecture*}{Conjecture}
\newtheoremstyle{nonindented}{1ex}{1ex}{}{}{\bfseries}{.}{.5em}{}
\newtheoremstyle{indented}{1ex}{1ex}{\itshape\addtolength{\leftskip}{0.6cm}\addtolength{\rightskip}{0.6cm}}{}{\bfseries}{.}{.5em}{}
\theoremstyle{nonindented}
\theoremstyle{indented}
\theoremstyle{plain}

\renewcommand{\hat}{\widehat}

\newcommand{\eat}[1]{}

\newenvironment{lp*}{\begin{equation*}  \begin{array}{lll}}{\end{array}\end{equation*}}

\AtBeginDocument{%
 }

\copyrightyear{2023}
\acmYear{2023}
\setcopyright{rightsretained}
\acmConference[KDD '23]{Proceedings of the 29th ACM SIGKDD Conference on Knowledge Discovery and Data Mining}{August 6--10, 2023}{Long Beach, CA, USA}
\acmBooktitle{Proceedings of the 29th ACM SIGKDD Conference on Knowledge Discovery and Data Mining (KDD '23), August 6--10, 2023, Long Beach, CA, USA}
\acmDOI{10.1145/3580305.3599763}
\acmISBN{979-8-4007-0103-0/23/08}

\makeatletter
\gdef\@copyrightpermission{
  \begin{minipage}{0.3\columnwidth}
   \href{https://creativecommons.org/licenses/by/4.0/}{\includegraphics[width=0.90\textwidth]{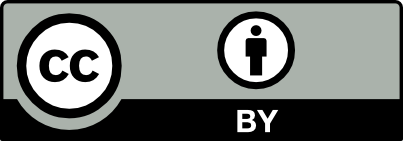}}
  \end{minipage}\hfill
  \begin{minipage}{0.7\columnwidth}
   \href{https://creativecommons.org/licenses/by/4.0/}{This work is licensed under a Creative Commons Attribution International 4.0 License.}
  \end{minipage}
}
\makeatother

\copyrightyear{2023}
\acmYear{2023}
\setcopyright{rightsretained}
\acmConference[KDD '23]{Proceedings of the 29th ACM SIGKDD Conference on Knowledge Discovery and Data Mining}{August 6--10, 2023}{Long Beach, CA, USA}
\acmBooktitle{Proceedings of the 29th ACM SIGKDD Conference on Knowledge Discovery and Data Mining (KDD '23), August 6--10, 2023, Long Beach, CA, USA}
\acmDOI{10.1145/3580305.3599763}
\acmISBN{979-8-4007-0103-0/23/08}
\newtheorem{assumption}{Assumption}

\begin{document}
\title{A Look into Causal Effects under Entangled Treatment in Graphs: \\ Investigating the Impact of Contact on MRSA Infection}

\author{Jing Ma}
\affiliation{%
  \institution{University of Virginia}
  \country{}
  }
\email{jm3mr@virginia.edu}

\author{Chen Chen}
\affiliation{%
  \institution{University of Virginia}
  \country{}
  }
\email{zrh6du@virginia.edu}

\author{Anil Vullikanti}
\affiliation{%
  \institution{University of Virginia}
  \country{}
  }
\email{vsakumar@virginia.edu}

\author{Ritwick	Mishra}
\affiliation{%
  \institution{University of Virginia}
  \country{}
  }
\email{mbc7bu@virginia.edu}

\author{Gregory	Madden}
\affiliation{%
  \institution{University of Virginia}
  \country{}
  }
\email{grm7q@hscmail.mcc.virginia.edu}

\author{Daniel Borrajo}
\affiliation{%
  \institution{J.P. Morgan AI Research}
  \country{}
  }
\email{daniel.borrajo@jpmchase.com}

\author{Jundong Li}
\affiliation{%
  \institution{University of Virginia}
  \country{}
  }
\email{jundong@virginia.edu}

\renewcommand{\shortauthors}{Jing Ma et al.}

\newcommand{\red}{\color{black}}
\newcommand{\indep}{\perp \!\!\! \perp}
\newcommand{\mymodel}{NEAT}
\newcommand{\bigCI}{\mathrel{\text{\scalebox{1.07}{$\perp\mkern-10mu\perp$}}}}

\begin{abstract}
Methicillin-resistant Staphylococcus aureus (MRSA) is a type of bacteria resistant to certain antibiotics, making it difficult to prevent MRSA infections. Among decades of efforts to conquer infectious diseases caused by MRSA, many studies have been proposed to estimate the causal effects of close contact (treatment) on MRSA infection (outcome) from observational data.
In this problem, the treatment assignment mechanism plays a key role as it determines the patterns of missing counterfactuals --- the fundamental challenge of causal effect estimation. Most existing observational studies for causal effect learning assume that the treatment is assigned individually for each unit. However, on many occasions, the treatments are pairwisely assigned for units that are connected in graphs, i.e., the treatments of different units are entangled. Neglecting the entangled treatments can impede the causal effect estimation. In this paper, we study the problem of causal effect estimation with treatment entangled in a graph. Despite a few explorations for entangled treatments, this problem still remains challenging due to the following challenges: (1) the entanglement brings difficulties in modeling and leveraging the unknown treatment assignment mechanism; (2) there may exist hidden confounders which lead to confounding biases in causal effect estimation; (3) the observational data is often time-varying. To tackle these challenges, we propose a novel method \mymodel, which explicitly leverages the graph structure to model the treatment assignment mechanism, and mitigates confounding biases based on the treatment assignment modeling. We also extend our method into a dynamic setting to handle time-varying observational data. Experiments on both synthetic datasets and a real-world MRSA dataset validate the effectiveness of the proposed method, and provide insights for future applications.
\end{abstract}



\maketitle

\section{Introduction}

{\red In the past a few decades, a burgeoning body of studies \cite{gurusamy2013antibiotic,kallen2010health,stefani2012meticillin} have been proposed for preventing infectious diseases such as Methicillin-resistant Staphylococcus aureus (MRSA). MRSA is a type of bacteria that is resistant to antibiotics, including methicillin and other penicillins. It can cause infections in the skin, respiratory tract, and urinary tract and can be spread through close contact with infected individuals or contaminated surfaces. In these scenarios, in-person contact relations are crucial for MRSA-related studies, and graphs are naturally used for modeling these relations. {\red An important question that medical specialists are interested in is: ``What is the causal effect of close contact (\emph{treatment}) on the spread of MRSA (\emph{outcome}) in a room-sharing network?" }}
Inspiringly, an emerging field that aims to investigate causal effects rather than the statistical correlations between variables in graph data has attracted arising attention recently \cite{guo2020learning,ma2021deconfounding}. 
In general, causal effect learning \cite{pearl2009causality,imbens2015causal} aims to estimate the causal effect of a certain treatment on an outcome for different units. On graph data, causal effect learning has great potential in many real-world applications such as epidemiology \cite{el2013network,ma2022assessing}. {\red The progress in this area provides us with effective tools for investigating contact impact on MRSA infection.}


As discussed in \cite{imbens2015causal}, the fundamental challenge of causal effect learning is data missing---only one potential outcome (the one that corresponds to the treatment assignment) can be observed for each unit. 
{\red For example, 
for a patient with frequent physical contact with others, the potential outcome for this individual with infrequent contact (i.e., counterfactual) is unavailable.}
As the treatment assignment mechanism (i.e., how the treatment is assigned to different units) determines which part of the data is missing, treatment assignment plays an essential role in causal studies.
Currently, most existing studies are based on the individualistic treatment assignment \cite{imbens2015causal}, where the treatment is assigned individually for each unit. However, in graphs, the treatment is often assigned in a pairwise manner to units that are connected. 
{\red For example, the in-room contact in a room-sharing network is often not individually applied to each person. Instead, it often happens between a pair of people.}
In these scenarios, treatments are not individually applied to each unit (i.e., treatments cannot be determined only by each unit's own properties). 
This setting is referred to as \textit{entangled treatment} \cite{toulis2018propensity}. 
Motivated by these scenarios, in this work, we study the problem of causal effect learning in graphs under entangled treatment. 

A few previous works \cite{toulis2018propensity,toulis2021estimating} have made preliminary explorations of this problem, but many challenges remain unaddressed: 
1) As discussed in \cite{toulis2018propensity}, treatment entanglement increases the risk of misspecification of the treatment effect estimator. 
If the entanglement through the graph is not considered, causal effect estimators tend to incorrectly attribute the observed treatment assignments to each unit's individual properties, and thus degrade the performance of causal effect estimation. 
To handle this entanglement problem, existing works \cite{toulis2018propensity} assume that the treatment assignment is determined by a pre-determined function over the graph (e.g., the treatment can be the node degree on the graph). However, on many occasions, this function is unknown.
2) Existing works \cite{toulis2018propensity,toulis2021estimating} rely on the unconfoundedness assumption \cite{rubin2005bayesian} (or its weaker version) that there do not exist unobserved confounders (confounders are variables which causally influence both the treatment and the outcome. 
{\red For example, patients' behavior habits are hidden confounders that influence their physical contact and infection risk. However, hidden confounders often exist in the real world and could lead to confounding biases.}
3) Existing works are often limited to a static setting. However, the graph, treatment, outcome, and unit covariates are naturally dynamic in many real-world scenarios. 
{\red For example,  the patient data is evolving over time; the causal association across different timestamps also brings more difficulties in learning causal effects.}
%


To address the aforementioned challenges, in this paper, we propose a novel framework \mymodel~ to estimate causal effects under \textbf{N}etwork \textbf{E}nt\textbf{A}ngled \textbf{T}reatments. Specifically: 
1) To handle the entangled treatment, for each node, we explicitly leverage its relevant graph topology to model the unknown treatment assignment with a learnable neural network module. 
2) To tackle the hidden confounders, we take the graph structure regarding each node as an instrumental variable (IV) \cite{hartford2017deep}. IV can eliminate the biases brought by hidden confounders in causal effect estimation.
{\red In the previous example, the room-sharing network is a valid IV if it is assumed to be independent of the patient's behavior habits, and its influence on the MRSA infection is fully mediated by the physical contact. }
A valid IV can provide a sort of randomization in the process of causal effect estimation and improve the estimation performance.
3) To learn causal effects in a dynamic setting, we generalize the setting and develop our framework to handle this problem across multiple timestamps. 

Notice that our work differs from other two areas of causal effect learning on graphs: 1) interference: these works \cite{ma2021causal,ma2022learning} assume that the treatment of each unit could causally affect the outcome of other units; 2) network deconfounding \cite{guo2020learning,ma2021deconfounding}: these works assume that hidden confounders  are buried in the graph structure. These two lines of work and our paper study separate research problems with different assumptions and application scenarios. In this work, our contributions can be summarized as follows: 
\begin{itemize}
    \item \textbf{Problem}. Motivated by the MRSA clinical studies, we investigate the important problem of causal effect estimation under entangled treatment. We address the challenges of treatment entanglement, hidden confounders, and a time-evolving environment. To the best of our knowledge, this is the first work addressing these challenges of this problem.
    \item \textbf{Method}. We propose a novel method, \mymodel, to address this problem. \mymodel~estimates causal effects with treatments entangled through a  graph. This method leverages the graph topology w.r.t. each node to better model the treatment assignment and facilitate treatment effect estimation even with hidden confounders. This method works for both static and dynamic settings. 
    \item \textbf{Experiment}. We conduct extensive experiments to evaluate our method on both synthetic and real-world graphs. Especially, we include real-world clinical data for MRSA infection. The results validate the effectiveness of our proposed method in different aspects. 
\end{itemize}

\section{Preliminaries}
\begin{figure}[t]
\centering
  \begin{subfigure}[b]{0.15\textwidth}
        \centering
        \includegraphics[height=1.1in]{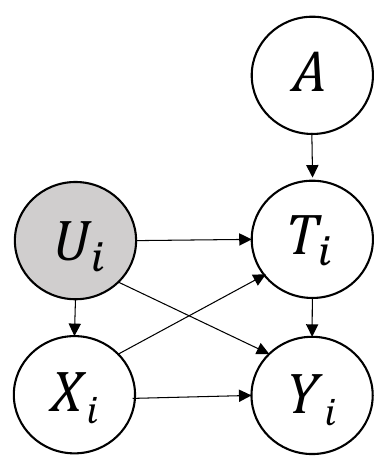}
        \caption{Static}
    \end{subfigure}
  \begin{subfigure}[b]{0.28\textwidth}
        \centering
        \includegraphics[height=1.1in]{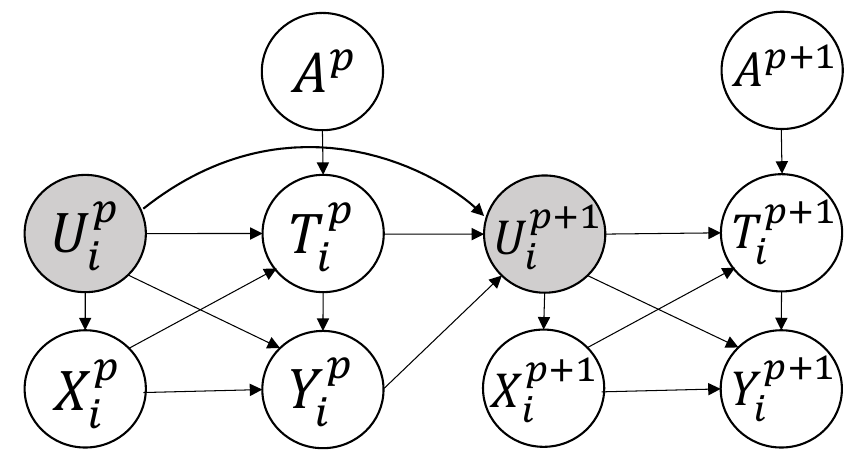}
        \caption{Dynamic}
    \end{subfigure}
  \caption{The causal graph of the studied problem in a static setting and in a dynamic setting. Each vertex in the causal graph represents a variable, and each arrow $A\rightarrow B$ represents a causal relation from $A$ to $B$. The observable variables are shown in white while the unobserved ones are shown in grey.}
  \label{fig:causal_graph}
\end{figure}

\subsection{Notations and Definitions}
The observational data is denoted by $\{\bm{X}, \bm{A}, \bm{T}, \bm{Y}\}^{1,...,P}$, which corresponds to the node features (e.g., patients' covariates), graph adjacency matrices (e.g., room-sharing network), treatment assignments (e.g., close contact), and observed outcomes (e.g., MRSA test result), respectively, in $P$ timestamps. We use $(\cdot)^p$ to denote the data in the $p$-th timestamp. When we focus on a static setting or a single timestamp, we drop this superscript for notation simplicity. We assume there are $N$ units (nodes) with $d_x$ covariates, with $\bm{X}^p=\{X_i^p\}_{i\in [N]}$, and for each unit $i$, $X_i\in \mathbb{R}^{d_x}$. The graph structure connecting these units at each timestamp is an $N\times N$ binary matrix $\bm{A}^p=\{A_{i,j}^p\}_{i,j\in [N]}$, where $A_{i,j}^p=1$ when there is an edge from node $i$ to node $j$, otherwise $A_{i,j}^p=0$. 
The treatment is $\bm{T}^p=\{T_i^p\}_{i\in [N]}$. In most studies, treatment is assumed to be a binary value, but in this work, we allow it to be a $d_t$-size vector 
(e.g., a vector that describes patients' close contact patterns).
The observed outcomes are denoted by $\bm{Y}^p=\{Y_i^p\}_{i\in [N]}$. For each unit $i$ at timestamp $p$, $Y_i^p\in \mathbb{R}$. 
In this paper, we use bold letters (e.g., $\bm{X}^p$) to denote variables for all units, and use unbold letters (e.g., ${X}_i^p$) to denote variables for a single unit. For simplicity, we use the same notation for both variables and data. The  subscript $(\cdot)_i$ denotes the index of a unit. If it is not necessary to emphasize the index of a specific unit, we drop the subscript to denote a random unit. The causal graph for this study is shown in Fig.~\ref{fig:causal_graph}; in this case, not all the confounders can be directly observed or measured, thus they can often lead to biased treatment effect estimation. The hidden confounders are denoted by $\bm{U}^p=\{U_i\}^p_{i\in [N]}$. 

This work is based on the well-known Neyman-Rubin potential outcome framework \cite{rubin2005causal}. The potential outcome is defined as the outcome which would have been realized when the treatment assignment had been set to a certain value. We denote the potential outcomes under treatment $T=t$ as $\bm{Y}^p(t)=\{Y_i(t)\}_{i\in [N]}^p$. Consider a baseline treatment as $T=t_0$, for a treatment $T=t$, the treatment effect conditioned on covariates $X$ in a static setting is defined as:
\begin{equation}
    \tau(X) = \mathbb{E}[Y_i(t)-Y_i(t_0)|X].
\end{equation}
In a dynamic setting, we denote the historical information before timestamp $p$ as $\bm{M}^{p}=\{\bm{X},\bm{A},\bm{T},\bm{Y}\}^{1,...,p-1}$. Similar to the above, we denote the historical information regarding unit $i$ before timestamp $p$ as $M_i^p$. When estimate causal effects at timestamp $p$, only the data no later than timestamp $p$ can be used. We define the treatment effect at timestamp $p$ as:
\begin{equation}
    \tau(X^p, \bm{M}^{p}) = \mathbb{E}[Y_i^p(t)-Y^p_i(t_0)|X^p, \bm{M}^{p}].
\end{equation}
Similar as \cite{shalit2017estimating}, we define the treatment effect for each unit $i$ at timestamp $p$ as $\tau^p_i=\tau(X^p_i, \bm{M}^{p})$. 

We define the entangled treatment as follows:
\begin{definition}
(Entangled treatment) The treatment here can be a function $\mathcal{T}(\cdot)$ over the graph structure, the observed features, and the hidden confounders: 
\begin{equation}
 T = \mathcal{T}(\bm{A}, {X}, {U}).
 \label{eq:treatment}
\end{equation}
In a dynamic setting, the treatment is also a function over historical information
\begin{equation}
    T^p = \mathcal{T}(\bm{A}^p, {X}^p, {U}^p, {M}^{p}).
    \label{eq:treatment_dynamic}
\end{equation}
\end{definition}

Notice that as the treatment function has the graph structure as an input, the treatments across different units are no longer individualistic (i.e., $T_i$ cannot be determined only based on variables of unit $i$). A typical example of the treatment function $\mathcal{T}(\cdot)$ is the degree of each node. But under many real-world circumstances, $\mathcal{T}(\cdot)$ is an unknown function.


The problem we study in this work is formally defined as:
\begin{definition} (Causal effect estimation under entangled treatments)
Given the observational data $\{\bm{X}, \bm{A}, \bm{T}, \bm{Y}\}^{1,...,P}$, we aim to estimate the treatment effect $\tau(X^p, \bm{M}^{p})$ for different units at each timestamp $p$ with treatments entangled in the graph.  
\end{definition}

\subsection{Assumptions}
\label{sec:assumption}

We assume that the outcome is generated by treatment, features, historical information, and hidden confounders as follows:
\begin{equation}
    Y^p = \mathcal{Y}(T^p, X^p, M^p) + g(U^p),
    \label{eq:counterfactual}
\end{equation}
where $\mathcal{Y}$ and $g$ are unknown and (nonlinear) functions. We assume $E[g(U^p)]=0$. In this work, we take the graph structure as an instrumental variable for IV analysis. An implicit assumption of our work is that the graph information of each node $i$ can be represented as a variable $A_i$, and its samples in observational data are sufficient for us to capture the patterns it influences the treatment assignment. The following assumptions make the graph structure as a valid IV.
\begin{assumption}
(Relevance) Given $X^p, M^p$ for any random unit, the treatment is relevant to the graph structure, i.e., $A^p\not\!\perp\!\!\!\perp T^p| X^p, M^p$.
\end{assumption}

\begin{assumption}
(Exclusion restriction) For any random unit, the causal effect of $A^p$ on $Y^p$ is fully mediated by $T^p$, i.e., $Y^p(T, A) = Y^p(T,A')$, $\forall A\ne A'$. Here, $Y^p(T, A)$ denotes the potential outcome for treatment $T$ and graph $A$ at timestamp $p$.
\end{assumption}

\begin{assumption}
(Instrumental unconfoundedness) There is no unblocked backdoor path from $A^p$ to $Y^p$, i.e., $A^p \!\perp\!\!\!\perp Y^p(A) | X^p, M^{p}$ for any random unit. Here, $Y^p(A)$ denotes the potential outcome for graph $A$ at timestamp $p$.
\end{assumption}

Inspired by recent IV studies \cite{hartford2017deep}, we use the above assumptions to effectively leverage the graph structure as an IV for  treatment effect estimation. More analysis can be found in Appendix A.

\begin{figure*}[t]
\centering
     \includegraphics[width=0.75\textwidth, height=3.3in]{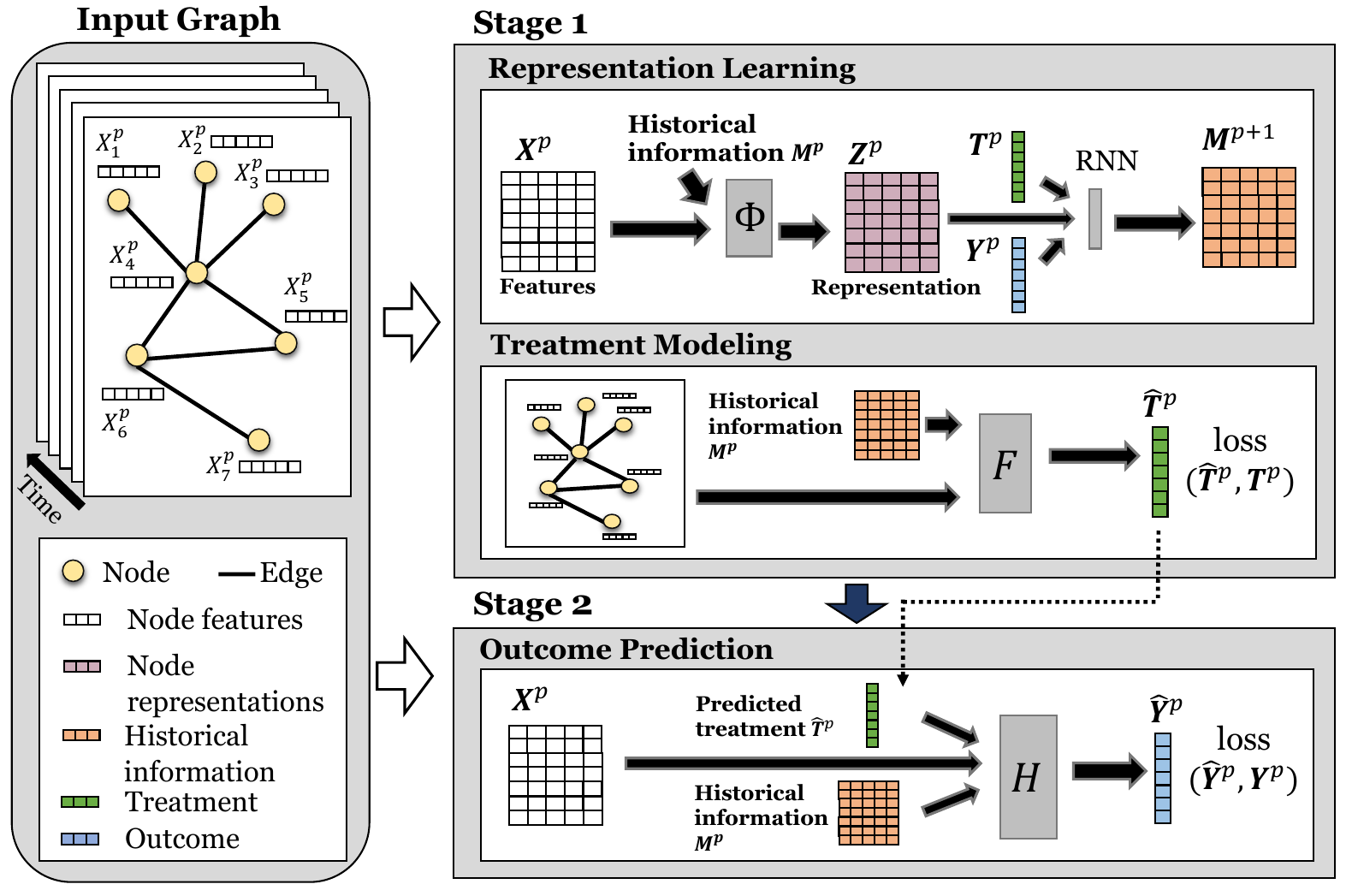}
      \caption{The proposed framework \mymodel. It contains three components: node representation learning, entangled treatment modeling, and outcome prediction.}
       \label{fig:framework}
\end{figure*}

\section{The Proposed Framework}
In this section, we introduce the proposed  \mymodel~ framework for causal effect learning under entangled treatment on the graph. Fig.~\ref{fig:framework} shows an illustration of the proposed framework. Specifically, this framework  contains three modules: node representation learning, entangled treatment modeling, and outcome prediction. 

\subsection{Overall Pipeline}
The whole framework is designed in a classical two-stage IV study pipeline \cite{angrist2009mostly,hartford2017deep}. 
Generally, in this pipeline, the first stage predicts the treatment with IVs, and the second stage estimates the potential outcomes based on the treatment predicted by the first stage. The key intuition behind this design is that, as the IVs are unconfounded, the predicted treatment from the first stage can provide more randomization, and thus it can help  mitigate the confounding bias brought by hidden confounders.

In our framework, in the first stage, we train a treatment modeling module to predict treatment assignments for each node at each timestamp. In this module, we leverage the graph structure as an IV, and utilize it to capture the patterns of entangled treatment in the graph. Simultaneously, we learn a representation for each node to encode its  properties, including its current features and historical information.  
In the second stage, we predict potential outcomes based on the original node features, the learned node representations, and the predicted treatment. In this two-stage IV framework, the biases brought by hidden confounders can be effectively eliminated. 


\subsection{Node Representation Learning}
The treatment effects are often different for nodes with different properties. For example, close contact may influence patients of different ages differently. 
To model such heterogeneity, we capture the properties of each node through node representation learning.
For each node $i$, we learn a representation $Z_i$ to encode its properties based on its node features $X_i$: 
\begin{equation}
    Z_i=\phi(X_i).
\end{equation}
Here, $\phi(\cdot)$ is implemented by a neural network module with learnable parameters. 

\noindent\textbf{Dynamic setting.} In a time-evolving environment, as illustrated in Fig. ~\ref{fig:causal_graph} (b), the current properties of each node can be influenced by the historical data in previous timestamps. To capture the time-evolving properties and model the causal mechanism in a dynamic setting, for each node $i$, we embed the historical information before each timestamp $p$ into a representation $M_i^p$ with a recurrent neural network (RNN) \cite{hochreiter1997long,cho2014learning}. $M_i^P$ is then incorporated into $Z_i^p$. At each timestamp, we update the historical embedding as:
\begin{equation}
M_i^{p}=\text{RNN}(M_i^{p-1}, T_i^{p-1}, Y_i^{p-1}, Z_i^{p-1}, X_i^{p-1}). 
\end{equation}
Here, we learn the representation for each node $i$ at timestamp $p$ with a transformation function $\Phi(\cdot)$:
\begin{equation}
    Z_i^p = \Phi(X_i^{p},M_i^{p}).
\end{equation}


\subsection{Entangled Treatment Modeling}
The treatment function $\mathcal{T}(\cdot)$ in Eq. (\ref{eq:treatment}) or Eq. (\ref{eq:treatment_dynamic}) is often not pre-determined. To better estimate treatment effects from observational data, we capture the treatment assignment patterns by training a module $F(\cdot)$ to model the conditional distribution of treatment ${T}_i^p$ given $\bm{A}^p,{X}_i^p, {M}_i^p$. The treatment modeling module is trained in the first stage together with node representation learning:
\begin{equation}
    \hat{T}_i^p=F(\bm{A}^p,X_i^p, M_i^p)=f(\bm{A}^p,\Phi(X_i^{p},M_i^{p})).
\end{equation}

\noindent\textbf{Treatment Entanglement.} As the treatments of different units are entangled through the graph structure, to effectively capture the patterns of  treatment assignment, we explicitly leverage the graph structure in the treatment modeling module. As a feasible implementation, we design this module $F(\cdot)$ based on graph neural networks (GNNs) \cite{kipf2016gcn,velivckovic2017graph}. Here we use one-layer graph convoluntional network (GCN) \cite{kipf2016gcn} to predict the treatment as follows:
\begin{equation}
\hat{T}_i^p =\sigma(\hat{\bm{A}}^p ([\bm{X}^p, \bm{Z}^p]) \bm{W}_0),
\end{equation}
where $\sigma(\cdot)$ is an activation function such as Softmax. $\hat{\bm{A}}^p$ is the normalized adjacency matrix calculated from the graph $\bm{A}^p$ beforehand with the renormalization trick \cite{kipf2016gcn}. Here $[\cdot, \cdot]$ stands for the concatenation operation. $\bm{W}_0$ denotes the parameters in GCNs.

\noindent\textbf{Loss for treatment modeling.} The loss for treatment prediction is denoted by $\mathcal{L}_t$. Generally, $\mathcal{L}_t$ is defined as:
\begin{equation}
\mathcal{L}_t = \sum_{p=1}^P\sum_{i=1}^N l_t(\hat{T}_i^p, T_i^p)=\sum_{p=1}^P\sum_{i=1}^N l_t(F(\bm{A}^p,X_i^p, M_i^p), T_i^p),
\end{equation}
where $l_t(\cdot)$ is a loss term to measure the prediction error of treatment modeling. 
Noticeably, in this work, we do not restrict the data type of treatment. To handle different types of treatment, we design a different implementation for this module. More specifically, for discrete treatments 
(e.g., whether a patient has frequent close contact),
we implement treatment prediction $F(\cdot)$ as a classification model with the cross-entropy loss function; for continuous  treatments 
(e.g., values that describe the patient's contact patterns)
we implement this module as a prediction task with mean square error (MSE) loss. 



\subsection{Outcome Prediction}
We train an outcome prediction module $H(\cdot)$ in the second stage, which predicts $Y_i^p$ based on $M_i^p, X_i^p$, and $\hat{T}_i^p$:
\begin{equation}
    \hat{Y}_i^p = \int H(\hat{T}_i^p, M_i^p, X_i^p) d F(\hat{T}_i^p|\bm{A}^p,X_i^p,M_i^p).
\end{equation}
We denote the loss function for outcome prediction by:
\begin{equation}
    \mathcal{L}_y=\sum_{p=1}^P\sum_{i=1}^N l_y(\hat{Y}_i^p,Y_i^p),
\end{equation}
where $l_y(\cdot)$ is a loss function (e.g., MSE) to measure the prediction error of the outcome. 
For each node $i$, the potential outcome w.r.t. treatment $T=t$ is predicted by $\hat{Y}_i(t)=H(t, M_i, X_i)$. We thereby estimate the treatment effect for each node $i$ as:
\begin{equation}
    \hat{\tau}_i = \hat{Y}_i(t)-\hat{Y}_i(t_0).
\end{equation}



\subsection{Implementation Details}
In node representation learning, we implement $\Phi(\cdot)$ with a multi-layer perceptron (MLP) and use a Gated Recurrent Unit (GRU) \cite{cho2014learning} for RNN. 
In entangled treatment modeling, we implement $F(\cdot)$ with a GCN layer. For discrete treatments, we use Softmax as the final layer, and take the output logits to model the probability of treatment values. For continuous treatments, we model them with a mixture of Gaussian distribution with component weights $w_k(\bm{A}^p, X^p, M^p)$ and parameters $(\mu_k(\bm{A}^p, X^p, M^p),\sigma_k(\bm{A}^p, X^p, M^p))$ for each component $k$. 
In outcome prediction, we use an MLP module to implement $H(\cdot)$, and use MSE loss for $l_y(\cdot)$. We use two optimizers to train the first and the second stage, respectively. 


\subsection{Discussion}
Many graph learning techniques (e.g., GCNs) mainly focus on local graph information (generally, $k$-layer GCNs can handle neighbors within $k$ hops), but if the treatment assignment is affected by a wider range on the graph (e.g., the length of the longest path which contains node $i$), it would be more difficult to capture and handle such information. However, it is worth noting that the proposed framework should not be limited to the specific implementation as introduced above. Instead, we can replace each component with a different implementation to achieve better specifications if relevant prior knowledge is given. 

\section{Experiments}
In this section, we validate the effectiveness of our proposed
method by conducting extensive evaluations. More specifically, our experiments are designed to answer the following research questions:
(1) \textbf{RQ1:} How does the proposed framework perform under treatment entanglement compared with state-of-the-art baselines? 
(2) \textbf{RQ2:} How does the proposed framework perform under different  levels of treatment entanglement and hidden confounders?
(3) \textbf{RQ3:} How does each component of the proposed framework contribute to the final treatment effect estimation? 
(4) \textbf{RQ4:} How does the proposed framework perform under different parameter settings?
    
\begin{table}[t]
  \caption{Detailed statistics of the datasets.}
  \label{tab:dataset}
  \def\arraystretch{1.15}%
  \begin{tabular}{lllll}
    \toprule
    Dataset & Random & Transaction & Social  & MRSA\\
    \midrule
    \# of nodes & $30,000$ & $186,509$ & $52,406$& $11,044$\\ 
    \# of edges & $208,193$ & $61,572$ & $107,394$& $31,403$\\
    \# of features &$32$ & $21$ & $16$ & $8$\\
     \# of timestamps &$12$ & $15$ & $10$ & $13$\\
  \bottomrule
\end{tabular}
\end{table}

\subsection{Dataset and Simulation}

In our experiment, we use four datasets with dynamic graph data, including synthetic, semi-synthetic, and real-world data. 
As it is notoriously hard to obtain the true causal models and counterfactuals from the real world, on the first three datasets, we follow regular practice to evaluate our method on data with simulated causal models.
Nevertheless, we encourage our simulation to be as close to reality as possible, thus, our synthetic and semi-synthetic datasets are based on graphs that are generated by real-world relational information and node features. 
Based on these graph data, we simulate the time-varying hidden confounders, treatment assignments, and outcomes.

\subsubsection{Simulation} 
\label{sec: simulation}
We describe the way we simulate different variables as follows. More details of simulation are in Appendix B.

\noindent\textbf{Hidden confounders.}
In a static setting, we simulate the hidden confounders as:
\begin{equation}
 {U}_{i} \sim \mathcal{N}(0, \mu \bm{I}).
\end{equation}
Here, $\bm{I}$ denotes an identity matrix of size $d_u$ (i.e., the dimension of hidden confounders). We set $\mu=20$ by default.

\noindent\textbf{Features.}
If the node features are available in the dataset, we directly use them. Otherwise, we simulate them by:
\begin{equation}
    {X}_{i} = \psi({U}_{i}) + \epsilon_x,
     \label{eq:feature}
\end{equation}
where $\psi(\cdot)$ is a linear function $\mathbb{R}^{d_u}\rightarrow \mathbb{R}^{d_x}$. Here, $d_x$ is the dimension of node features. $\epsilon_x$ is a noise vector in Gaussian distribution.

\noindent\textbf{Treatment.}
We simulate the treatment with function $\mathcal{T}$:
\begin{equation}
    T_i = \texttt{BI}((1-\lambda){\Theta}_{t,x}^\top X_i + \lambda\frac{1}{|\mathcal{N}_i|}\sum_{j\in \mathcal{N}_i}(\Theta_{t,x}^\top X_j) + \Theta_{t,u}^\top U_i + \epsilon_t),
     \label{eq:treatment_sim}
\end{equation}
where ${\Theta}_{t,x},{\Theta}_{t,u}$ are parameter vectors with dimension $d_x$ and $d_u$, respectively. Each parameter in ${\Theta}_{t,*}$ is in Gaussian distribution $\mathcal{N}(0,0.5^2)$. $\mathcal{N}_i$ is the set of neighbors of node $i$ in the graph. We use only one-hop neighbors by default. $\lambda\in[0,1]$ is the parameter that controls the strength of treatment entanglement, i.e., the larger $\lambda$ is set, the stronger the graph influences the treatment assignments. $\texttt{BI}(\cdot)$ is a function that maps the input to a binary value. A regular implementation is to transform the input to a probability using a Sigmoid function, and then sample the output with Bernoulli distribution. Noticeably, we do not restrict the treatment to be a binary value. Continuous treatment can be simulated without the $\texttt{BI}(\cdot)$ function; and high-dimensional treatment with dimension $d_t$ can be simulated by replacing the parameter vector $\Theta_{t,x}$ with a parameter matrix ${\Theta}_{t,x}$ with dimension $d_x\times d_t$ (similarly for $\Theta_{t,u}$). $\epsilon_t\sim \mathcal{N}(0,0.01^2)$ is a random Gaussian noise.

\noindent\textbf{Potential outcome.}
We simulate the potential outcomes as follows:
\begin{equation}
    Y_i(t) = t \cdot \Theta_y^\top X_i + \Theta_0^\top  X_i + \beta \Theta_u^\top U_i + \epsilon_y,
     \label{eq:outcome}
\end{equation}
where $\Theta_y$ and $\Theta_0$ are parameter vectors of dimension $d_x$, and $\Theta_u$ is of dimension $d_u$. $\beta\ge0$ is a parameter that controls the strength of the hidden confounder. $\epsilon_y\sim \mathcal{N}(0,0.1^2)$ is a noise.

\noindent\textbf{Dynamic setting.} In a dynamic setting, we simulate the historical data and hidden confounders over time as:
\begin{equation}
     {M}_{i}^p = \sum_{r=1}^R (W_u^{r}{U}_{i}^{R-r} + W_x^{r}{X}_{i}^{R-r} +W_t^{r}{T}_i^{R-r} + W_y^{r}{Y}_i^{R-r}),
    \label{eq:history}
\end{equation}
\begin{equation}
    {U}_{i}^p = \psi_u ({M}_{i}^p) + \epsilon_u
\end{equation}
where $R$ is the number of previous timestamps which influence the current one. We set $R=3$ by default. Generally, the historical information at each timestamp encodes the previous hidden confounders, node features, treatments, and outcomes. Parameters $W_u^{r}$, $W_x^{r}$, $W_t^{r}$, and $W_y^{r}$ control these four types of influence from timestamp $R-r$. 
We generate time-varying hidden confounders with a transformation over the historical information. Here,  $\psi_u(\cdot)$ is a linear transformation function. $\epsilon_u\sim\mathcal{N}(0,\bm{I})$ is a noise. We use the same way as Eq. (\ref{eq:feature}) to simulate features. The treatments and outcomes are also generated similarly as above description in Eq. (\ref{eq:treatment_sim})  and Eq. (\ref{eq:outcome}), but the historical information ${M}_{i}^p$ is incorporated by concatenating it with $X_{i}^p$ as input.

\begin{table*}[t]
\centering
 \caption{Performance of treatment effect estimation for different methods.}
\def\arraystretch{1.1}%
  \begin{tabular}{l|cccccc|cccccc}
    \hline
    &\multicolumn{6}{c|}{Static}  & \multicolumn{6}{c}{Dynamic}  \\
    \cline{2-13}
    \multirow{2}{*}{Method} & \multicolumn{2}{c}{Random} & \multicolumn{2}{c}{Transaction} & \multicolumn{2}{c|}{Social}  & \multicolumn{2}{c}{Random} & \multicolumn{2}{c}{Transaction} & \multicolumn{2}{c}{Social} \\
    \cline{2-13}
    & $\sqrt{\epsilon_{PEHE}}$ & ${\epsilon_{ATE}}$ & $\sqrt{\epsilon_{PEHE}}$ & ${\epsilon_{ATE}}$ &  $\sqrt{\epsilon_{PEHE}}$ & ${\epsilon_{ATE}}$ &$\sqrt{\epsilon_{PEHE}}$ & ${\epsilon_{ATE}}$ & $\sqrt{\epsilon_{PEHE}}$ & ${\epsilon_{ATE}}$ &  $\sqrt{\epsilon_{PEHE}}$ & ${\epsilon_{ATE}}$ \\
    \hline
    SL & $67.2$\scriptsize{~$\pm3.0$} & $7.3$\scriptsize{~$\pm0.5$} & $40.9$\scriptsize{~$\pm1.4$} & $7.1$\scriptsize{~$\pm0.3$}& $48.3$\scriptsize{~$\pm2.5$} & $9.2$\scriptsize{~$\pm0.7$}& $69.4$\scriptsize{~$\pm3.1$} & $7.7$\scriptsize{~$\pm0.4$}& $55.8$\scriptsize{~$\pm1.8$} & $8.4$\scriptsize{~$\pm0.6$}& $45.3$\scriptsize{~$\pm1.4$} & $6.5$\scriptsize{~$\pm0.3$} \\
    CF & $33.7$\scriptsize{~$\pm2.1$} & $7.0$\scriptsize{~$\pm0.2$} & $30.9$\scriptsize{~$\pm1.8$} & $6.9$\scriptsize{~$\pm0.3$}& $23.6$\scriptsize{~$\pm1.1$} & $5.9$\scriptsize{~$\pm0.4$}& $36.2$\scriptsize{~$\pm2.4$} & $7.4$\scriptsize{~$\pm0.6$}& $39.6$\scriptsize{~$\pm1.2$} & $6.2$\scriptsize{~$\pm0.4$}& $31.4$\scriptsize{~$\pm1.0$} & $5.8$\scriptsize{~$\pm0.4$} \\ 
    CFR & $28.1$\scriptsize{~$\pm2.4$} & $6.3$\scriptsize{~$\pm0.5$}& $34.4$\scriptsize{~$\pm2.3$} & $5.6$\scriptsize{~$\pm0.9$}& $27.3$\scriptsize{~$\pm2.0$} & $5.2$\scriptsize{~$\pm0.5$}& $33.3$\scriptsize{~$\pm2.7$} & $6.7$\scriptsize{~$\pm0.4$}& $30.0$\scriptsize{~$\pm2.6$} & $5.9$\scriptsize{~$\pm0.4$}& $27.7$\scriptsize{~$\pm2.2$} & $6.0$\scriptsize{~$\pm0.5$}\\
    \hline
    NetDeconf & $35.6$\scriptsize{~$\pm3.0$} & $6.2$\scriptsize{~$\pm0.3$}& $28.6$\scriptsize{~$\pm2.0$} & $5.8$\scriptsize{~$\pm0.7$}& $30.5$\scriptsize{~$\pm2.7$} & $6.3$\scriptsize{~$\pm0.4$}& $34.0$\scriptsize{~$\pm2.5$} & $6.8$\scriptsize{~$\pm0.7$}& $29.4$\scriptsize{~$\pm1.5$} & $6.1$\scriptsize{~$\pm0.5$}& $32.9$\scriptsize{~$\pm2.2$} & $5.8$\scriptsize{~$\pm0.8$}\\
    DNDC & $32.9$\scriptsize{~$\pm2.4$} & $6.8$\scriptsize{~$\pm0.3$}& $29.8$\scriptsize{~$\pm2.2$} & $6.0$\scriptsize{~$\pm0.5$}& $33.2$\scriptsize{~$\pm3.1$} & $6.6$\scriptsize{~$\pm0.7$}& $29.9$\scriptsize{~$\pm2.2$} & $6.2$\scriptsize{~$\pm0.4$}& $28.9$\scriptsize{~$\pm1.8$} & $5.7$\scriptsize{~$\pm0.5$}& $35.8$\scriptsize{~$\pm3.0$} & $6.4$\scriptsize{~$\pm0.7$}\\
    \hline 
    DeepIV & $31.0$\scriptsize{~$\pm2.3$}  & $5.9$\scriptsize{~$\pm0.4$}& $26.7$\scriptsize{~$\pm1.9$} & $5.4$\scriptsize{~$\pm0.6$}& $21.4$\scriptsize{~$\pm1.6$} & $5.1$\scriptsize{~$\pm0.3$}& $32.2$\scriptsize{~$\pm3.1$}  & $5.8$\scriptsize{~$\pm0.5$}& $30.2$\scriptsize{~$\pm1.9$} & $5.8$\scriptsize{~$\pm0.4$}& $24.1$\scriptsize{~$\pm1.8$} & $5.6$\scriptsize{~$\pm0.6$}\\
    \hline 
    \textbf{\mymodel} & $\bm{22.4}$\scriptsize{~$\bm{\pm1.8}$} & $\bm{5.2}$\scriptsize{~$\bm{\pm0.3}$}& $\bm{18.8}$\scriptsize{~$\bm{\pm1.4}$} & $\bm{4.6}$\scriptsize{~$\bm{\pm0.4}$}& $\bm{17.9}$\scriptsize{~$\bm{\pm1.2}$} & $\bm{4.1}$\scriptsize{~$\bm{\pm0.5}$}& $\bm{20.1}$\scriptsize{~$\bm{\pm1.4}$} & $\bm{5.0}$\scriptsize{~$\bm{\pm0.2}$}& $\bm{22.5}$\scriptsize{~$\bm{\pm1.0}$} & $\bm{5.3}$\scriptsize{~$\bm{\pm0.3}$}& $\bm{18.2}$\scriptsize{~$\bm{\pm1.6}$} & $\bm{5.0}$\scriptsize{~$\bm{\pm0.4}$}\\
    \hline
\end{tabular}
\vspace{3mm}
 \label{tab:baseline}
\end{table*}

\subsubsection{Datasets} We further introduce more details about each dataset used in this paper. More details of data statistics are shown in Table \ref{tab:dataset}, including the number of nodes, edges, features, and timestamps.

\noindent\textbf{Random graph.} This dataset contains synthetic graphs generated by the Erd\"{o}s-R\'{e}nyi (E-R) model \cite{erdHos1960evolution} at each timestamp. We use NetworkX \cite{hagberg2008exploring} to generate these graphs. Based on these graphs, we simulate other variables as described in Section \ref{sec: simulation}.

{\red\noindent\textbf{Real-world graphs.} We use two real-world dynamic graphs with each node representing a real person and each edge representing a certain type of connection between them. Based on the type of connection, these two datasets are referred as \textbf{Transaction} and \textbf{Social}, respectively. We use the covariates of people in these datasets as node features, and simulate the treatments and outcomes as described in Section \ref{sec: simulation}.} More details of these datasets can be found in Appendix B.

\noindent\textbf{MRSA.} This dataset contains real-world hospital data for studying Methicillin-resistant Staphylococcus aureus (MRSA) infection. We construct a dynamic graph for the room-sharing relations between patients. At each timestamp, each node is a patient, and an edge exists between a pair of patients if and only if they have shared at least one room during this timestamp. 
The patient information such as medicine usage and length of stay are taken as node features. We investigate the causal effect of the number of in-room contacts (treatment) on MRSA infection test results (outcome). 
We consider there exist hidden confounders such as patients' behavior habits. In this dataset, we do not use any simulated data, and do not evaluate our causal effect estimation based on simulated counterfactuals. Instead, we use the domain knowledge regarding MRSA to confirm our findings.


\subsection{Baselines}
In the experiments, we compare our method with some state-of-the-art baselines. These baselines can be divided into the following three main categories:
\begin{itemize}
    \item \textbf{Individual units.} These methods are based on the assumption that different units are independent. They estimate the treatment effect by adjusting for confounders based on unit covariates. We adopt the widely-used methods including S-Learner (\textbf{SL}) \cite{kunzel2019metalearners}, causal forest (\textbf{CF}) \cite{causalForest}, and counterfactual regression (\textbf{CFR}) \cite{CFR}.
    \item \textbf{Network deconfounder.} These methods assume that there is a graph connecting different units. They mitigate confounding biases by using the graph structure as a proxy for hidden confounders. We use the network deconfounder (\textbf{NetDeconf}) \cite{guo2020learning} and the dynamic network deconfounder (\textbf{DNDC}) \cite{ma2021deconfounding}.
    \item \textbf{DeepIV.} This method \cite{hartford2017deep} uses instrumental variables to mitigate the confounding biases. For each node $i$, we take the $i$-th row in the adjacency matrix as its IV.
\end{itemize}
We use the implementation released in the EconML package\footnote{https://github.com/microsoft/EconML} for S-Learner, causal forest, and DeepIV. 

\subsection{Evaluation Metrics}
We adopt two widely-adopted metrics for treatment effect estimation, including Rooted Precision in Estimation of Heterogeneous Effect (PEHE) \cite{BART} and Mean Absolute Error (ATE) \cite{willmott2005advantages} at each timestamp $p$:
\begin{equation}
    \sqrt{\epsilon_{PEHE}^p} = \sqrt{\frac{1}{N}\sum\nolimits_{i\in [N]}(\hat{\tau_i}^p-\tau_i^p)^2},
\end{equation}
\begin{equation}
    \epsilon_{ATE}^p = |\frac{1}{N}\sum\nolimits_{i\in [N]}\hat{\tau_i}^p-\frac{1}{N}\sum\nolimits_{i\in [N]}\tau_i^p|.
\end{equation}

For all the experiments, we calculate the average values of these metrics over all timestamps, and still denote them by $\sqrt{\epsilon_{PEHE}}$ and $\epsilon_{ATE}$ for simplicity. 

\subsection{Setup}
For all datasets, we randomly split them into $60\%/20\%/20\%$
training/validation/test data. By default, we focus on the dynamic setting and set the number of training epochs as 2000, the learning rate as $0.004$, the dimension
for node representation and history embedding as 32 and 20, respectively, $\lambda=0.5,\beta=0.5$. We report the mean and standard deviation of performance over ten repeated executions on test data. More details of experiment setup are in Appendix B.

\subsection{RQ1: Performance of Different Methods}
To demonstrate the effectiveness of the proposed method, in Table \ref{tab:baseline}, we show the treatment effect estimation performance of our method and the baselines in both static and dynamic settings. We observe that in both settings, the proposed method \mymodel~ outperforms other baselines in different metrics. We attribute the improvement to two key factors: 
1) We explicitly incorporate the graph structure to model the treatment assignment. During this process, we can better utilize the observational data for treatment effect estimation. Among the baselines, SL, CF, and CFR do not consider the graph which connects different units; NetDeconf and DNDC can leverage graph structure, but they use the graph as a proxy to infer the hidden confounders. These methods, however, do not fit in well in the problem setting studied in this paper. 
2) We utilize the graph structure as an instrumental variable to eliminate the confounding biases. Among the baselines, SL, CF, and CFR are based on the unconfoundedness assumption; NetDeconf and DNDC assume the hidden confounders can be inferred from the graph structure. These assumptions cannot be satisfied in our datasets. DeepIV also takes the graph information as an instrumental variable to handle hidden confounders, but its performance is impeded due to the lack of proper techniques to handle graph data. 

\begin{figure}[t]
\centering
  \begin{subfigure}[b]{0.237\textwidth}
        \centering
        \includegraphics[height=1.05in]{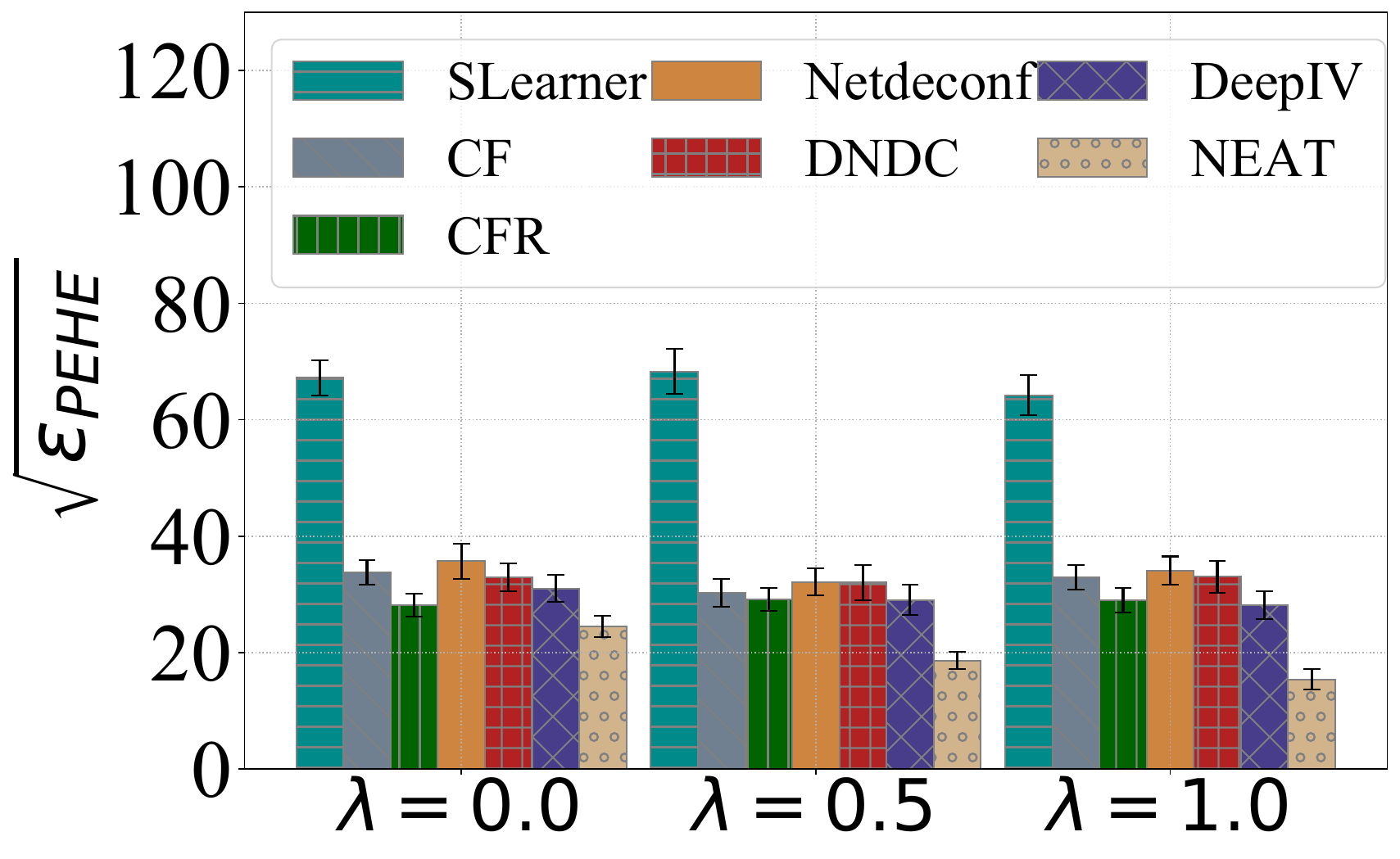}
        \caption{$\sqrt{\epsilon_{PEHE}}$}
    \end{subfigure}
  \begin{subfigure}[b]{0.235\textwidth}
        \centering
        \includegraphics[height=1.05in]{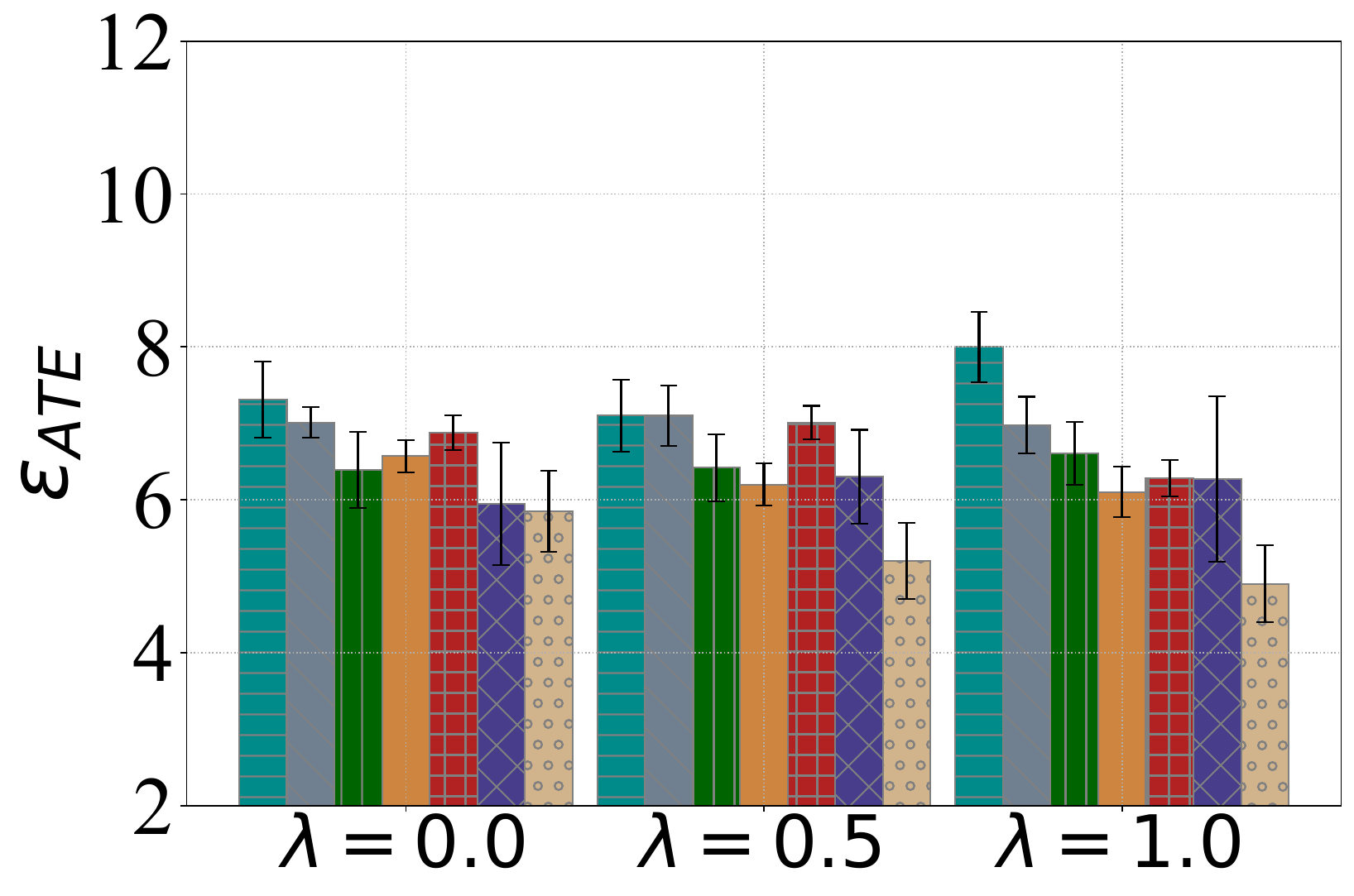}
        \caption{$\epsilon_{ATE}$}
    \end{subfigure}
  \caption{Treatment effect estimation performance under different levels of treatment entanglement on Random dataset.}
  \label{fig:lambda}
\end{figure}

\begin{figure}[t]
\centering
  \begin{subfigure}[b]{0.237\textwidth}
        \centering
        \includegraphics[height=1.05in]{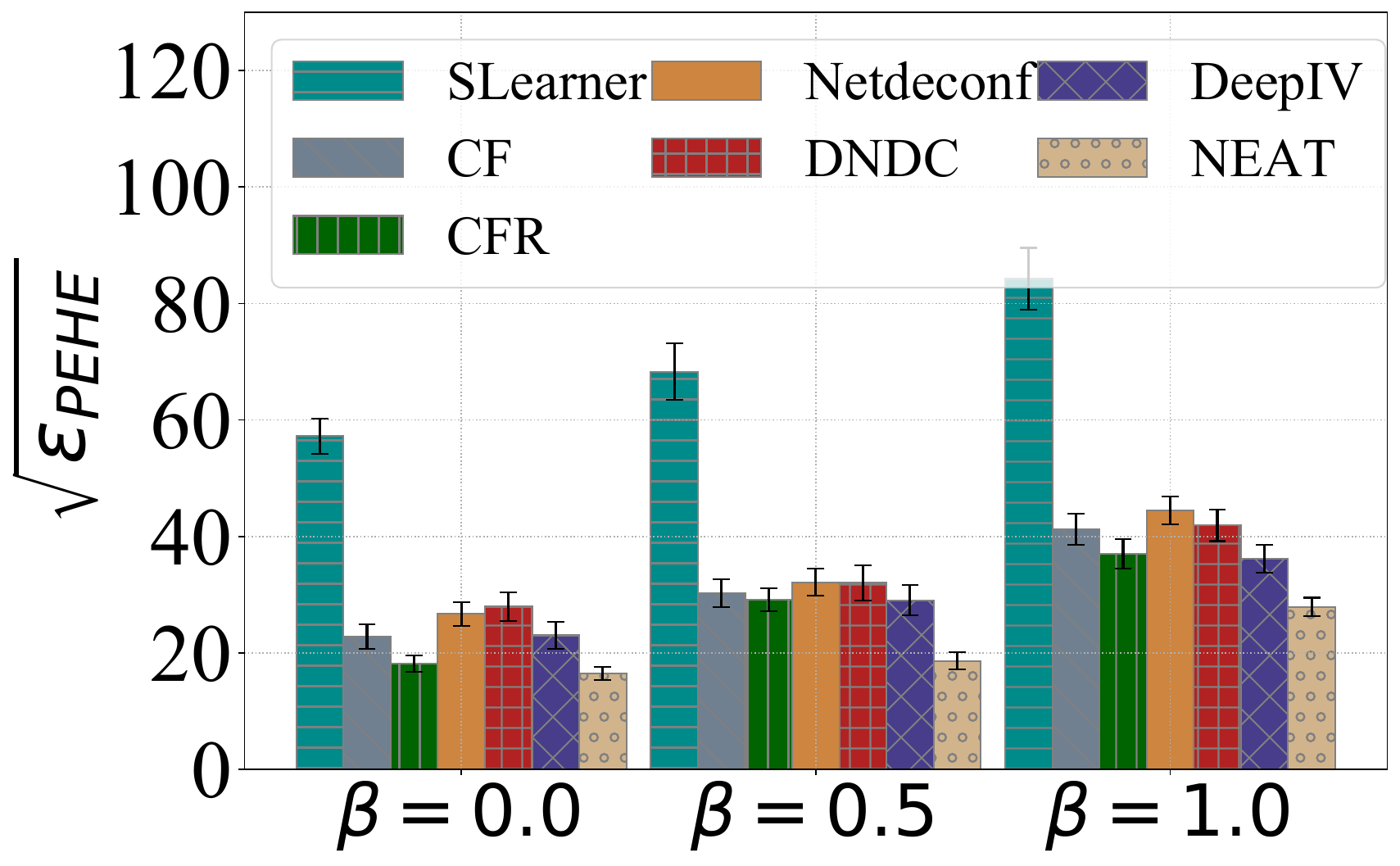}
        \caption{$\sqrt{\epsilon_{PEHE}}$}
    \end{subfigure}
  \begin{subfigure}[b]{0.235\textwidth}
        \centering
        \includegraphics[height=1.05in]{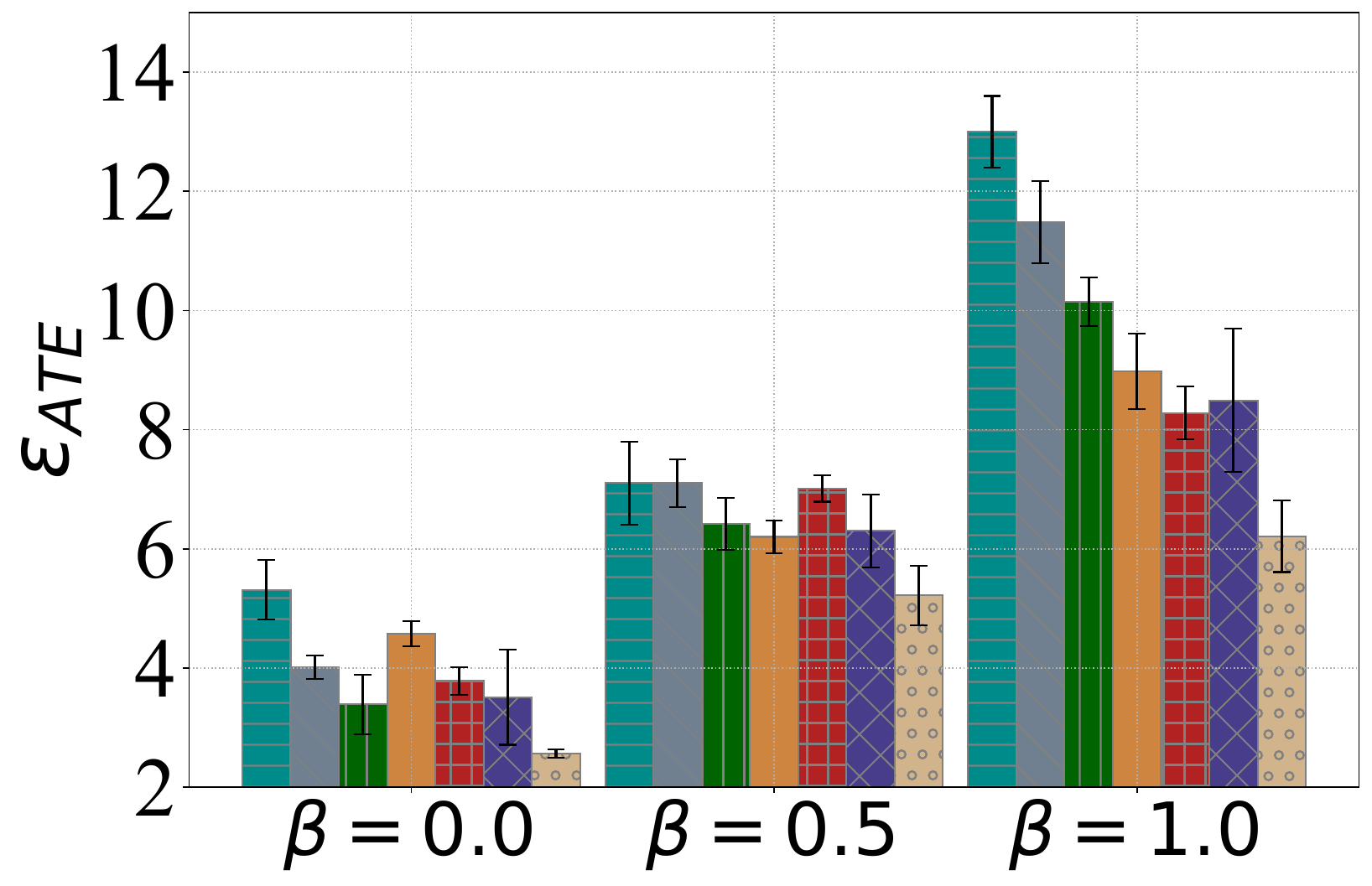}
        \caption{$\epsilon_{ATE}$}
    \end{subfigure}
  \caption{Treatment effect estimation performance under different levels of hidden confounders on Random dataset.}
  \label{fig:confounder}
\end{figure}

\begin{figure}[t]
\centering
  \begin{subfigure}[b]{0.23\textwidth}
        \centering
        \includegraphics[height=1.05in]{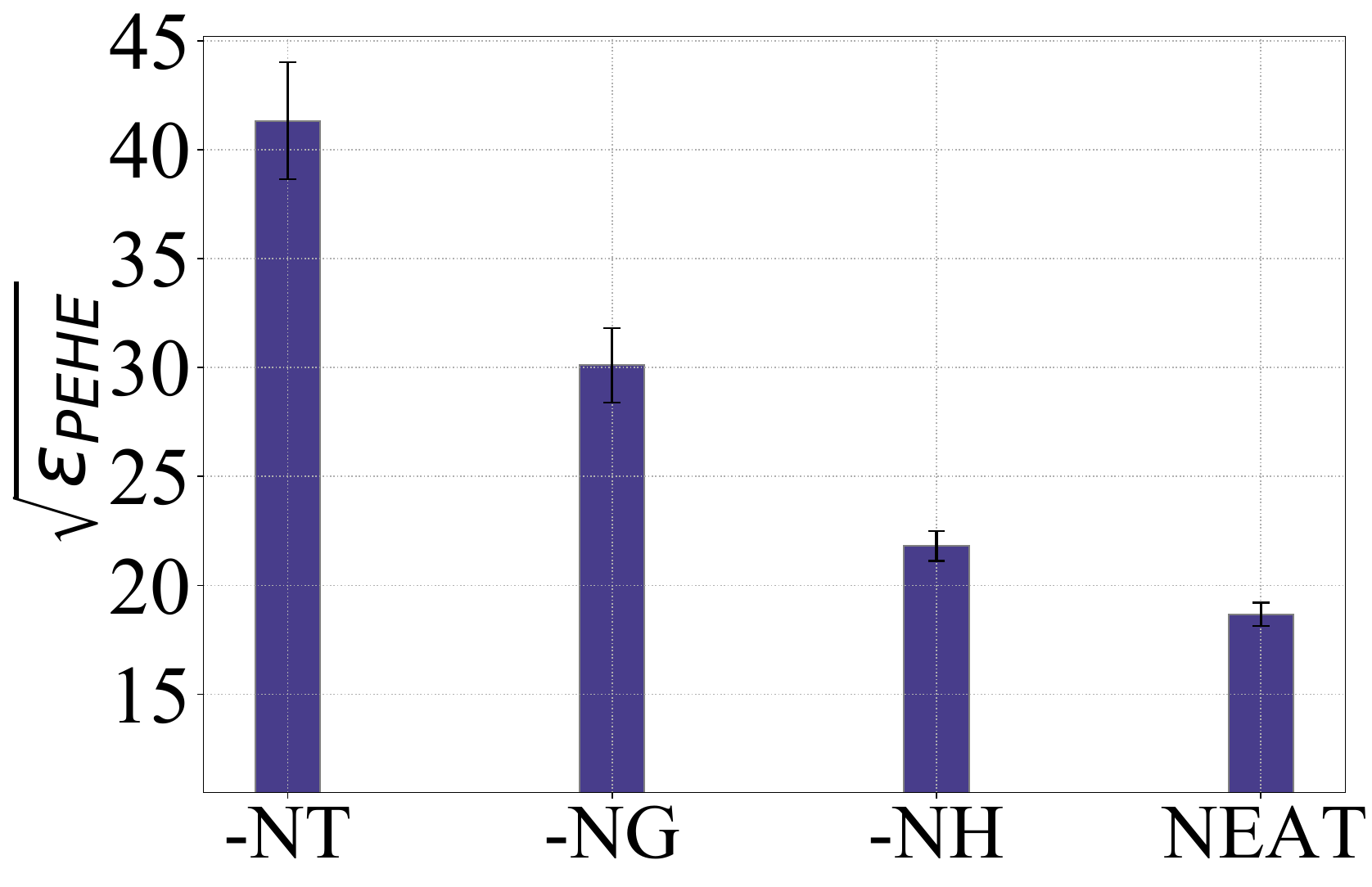}
        \caption{$\sqrt{\epsilon_{PEHE}}$}
    \end{subfigure}
  \begin{subfigure}[b]{0.23\textwidth}
        \centering
        \includegraphics[height=1.05in]{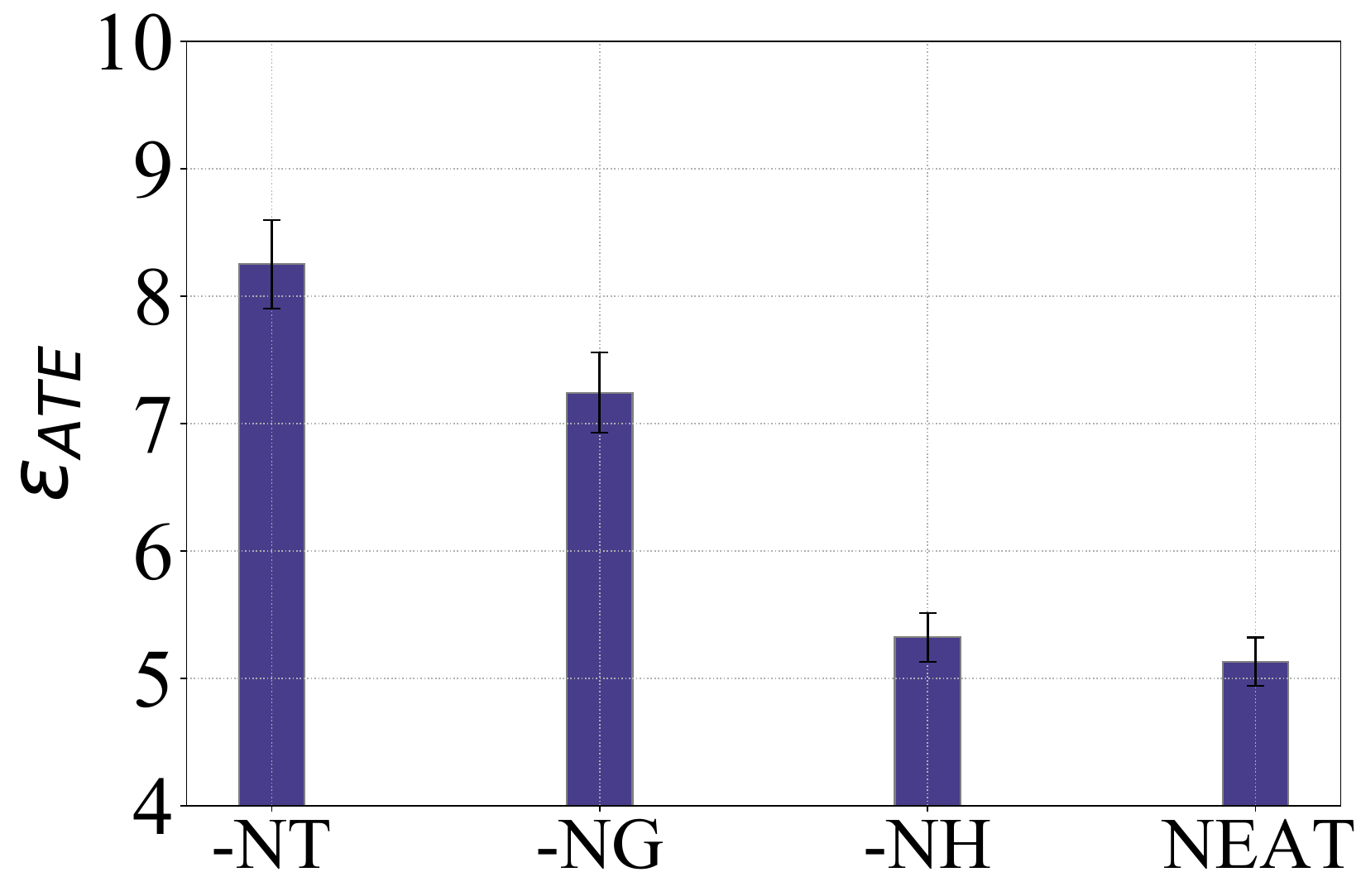}
        \caption{$\epsilon_{ATE}$}
    \end{subfigure}
  \caption{Ablation study for different variants of \mymodel~ on Random dataset.}
  \label{fig:ablation}
\end{figure}

\begin{figure}[t]
\centering
  \begin{subfigure}[b]{0.23\textwidth}
        \centering
        \includegraphics[height=1.12in]{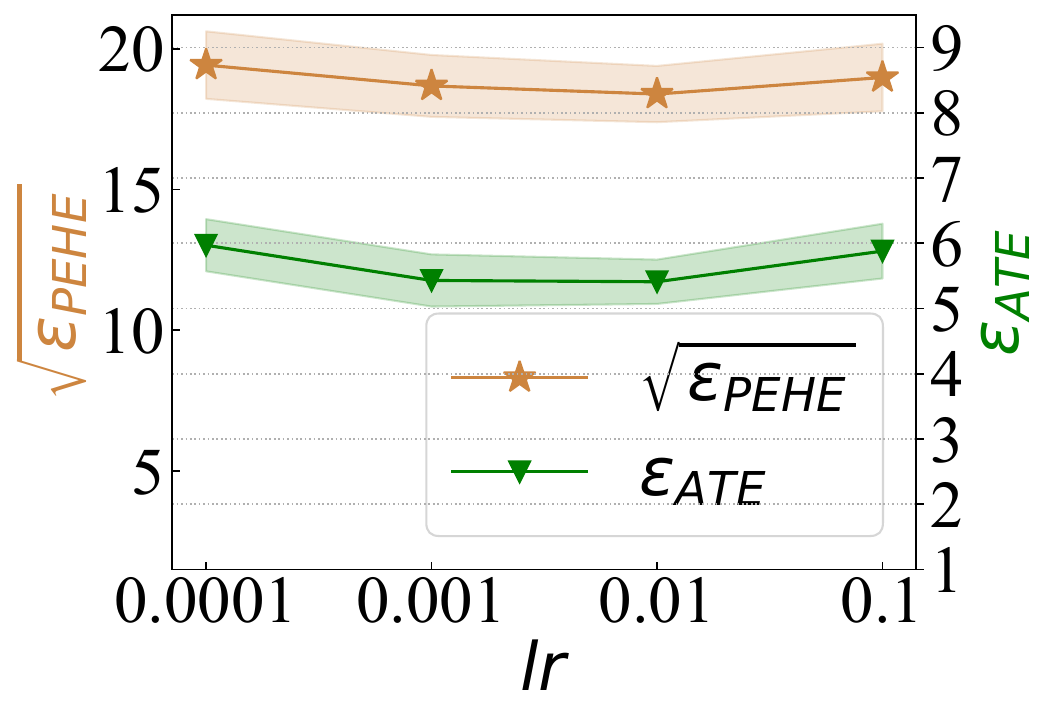}
        \caption{Learning rate}
    \end{subfigure}
  \begin{subfigure}[b]{0.23\textwidth}
        \centering
        \includegraphics[height=1.12in]{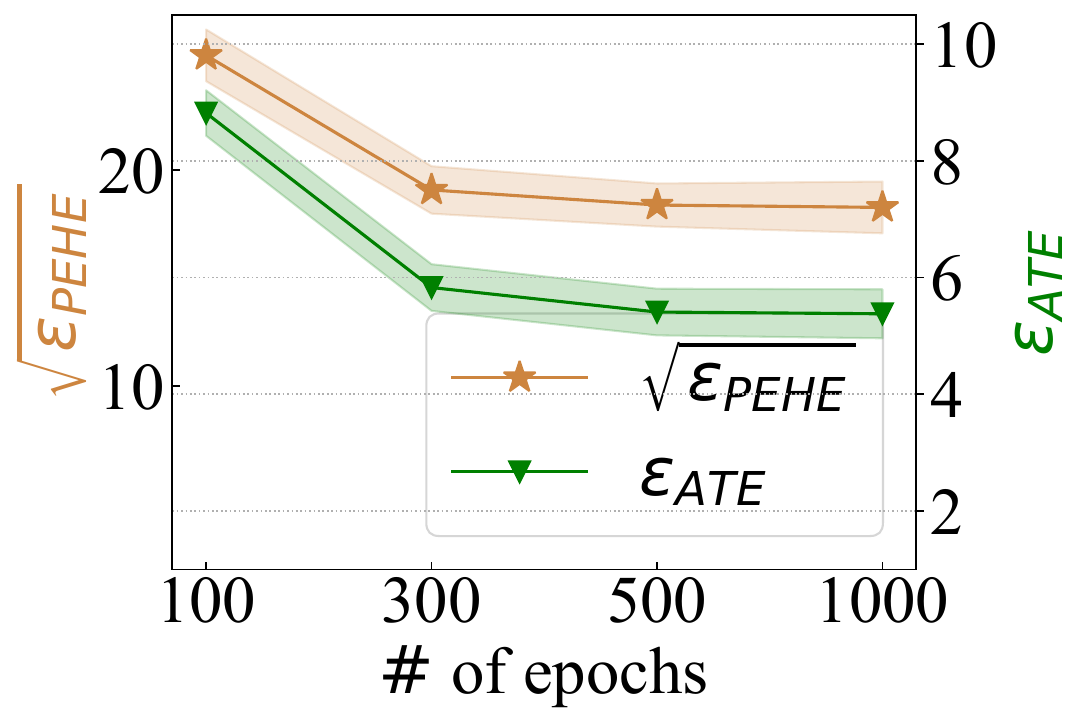}
        \caption{$\#$ of epochs}
    \end{subfigure}
    \begin{subfigure}[b]{0.23\textwidth}
        \centering
        \includegraphics[height=1.12in]{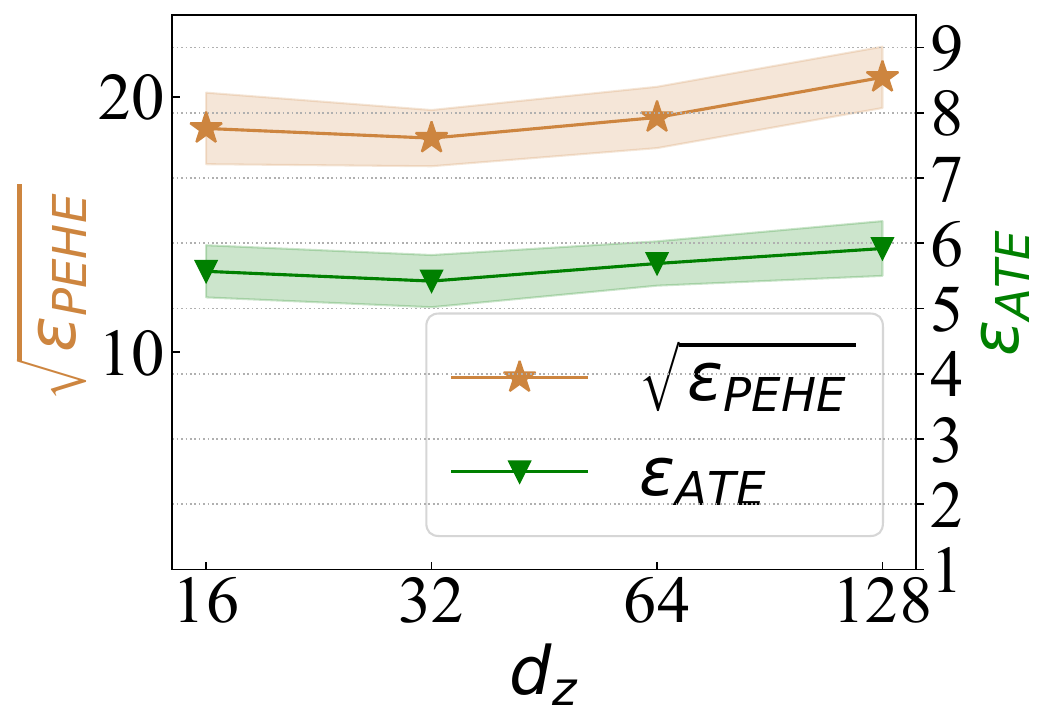}
        \caption{$d_z$}
    \end{subfigure}
  \begin{subfigure}[b]{0.23\textwidth}
        \centering
        \includegraphics[height=1.12in]{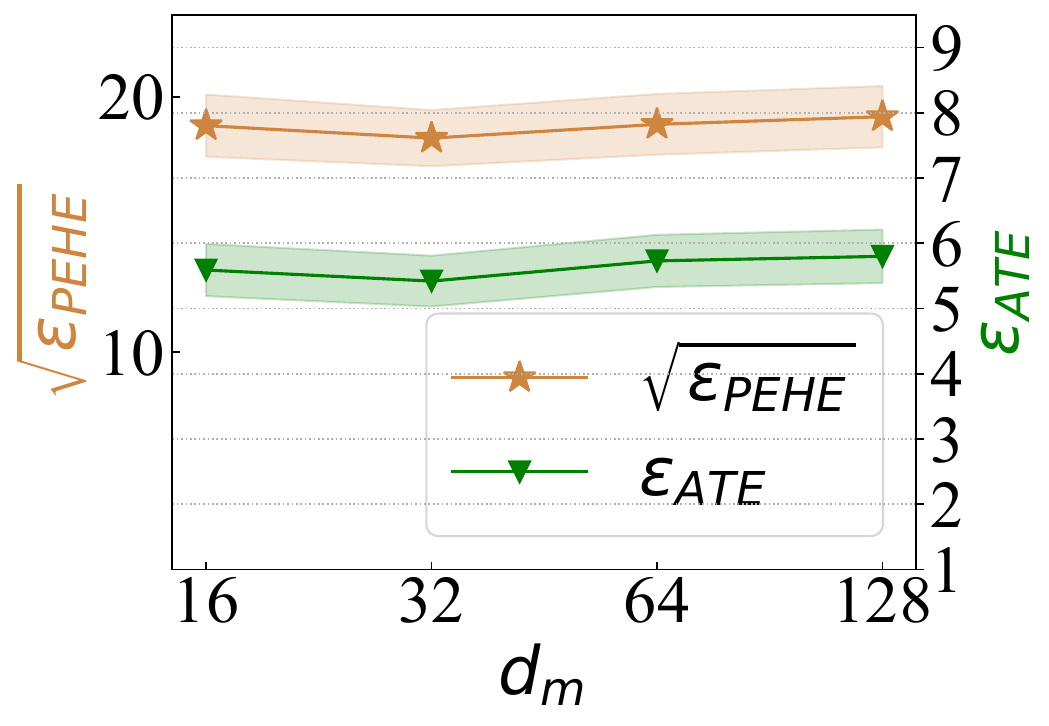}
        \caption{$d_m$}
    \end{subfigure}
  \caption{Parameter study for \mymodel~on Random dataset.}
  \label{fig:parameter}
\end{figure}

\subsection{RQ2: Performance under Different Levels of Treatment Entanglement and Confounders}

To evaluate our method more comprehensively, we test it under different levels of treatment entanglement. In the simulation, we control the treatment entanglement with parameter $\lambda$: the larger $\lambda$ is set, the stronger the treatment assignment of each node is entangled with neighbors. Fig.~\ref{fig:lambda} shows the causal effect estimation performance when we set $\lambda$ as different values. Generally, we observe more obvious performance gain when $\lambda$ is larger. This observation indicates that our method can well handle the entangled treatments by leveraging the graph structure. We only show the results on the Random dataset, but similar observations can also be found on other datasets.

We also evaluate our method under different levels of hidden confounders. In Fig.~\ref{fig:confounder}, we show the results when we change the strength of hidden confounders. Specifically, we change the strength by multiplying the hidden confounders in simulation with the parameter $\beta \ge 0$. From Fig.~\ref{fig:confounder}, it can be observed that compared with baselines, our method is more robust with hidden confounders. This is because we effectively utilize the graph as an instrumental variable to mitigate confounding biases.

\subsection{RQ3: Ablation Study}
To verify the effectiveness of each component in our method, we conduct an ablation study including the following variants: 
(1) \textbf{\mymodel-NT}: In this variant, we replace the treatment modeling module with a random sampling over the space of treatment assignment;
(2) \textbf{\mymodel-NG}: In this variant, we do not use the graph in treatment modeling, and replace the input adjacency matrix with an identity matrix.
(3) \textbf{\mymodel-NH}: In this variant, we remove the RNN in our method and do not use historical information. 
Fig.~\ref{fig:ablation} reports the performance of our method and these variants. The results show that all the different components contribute to the final superior performance of our method.

\subsection{RQ4: Parameter Study}
To investigate the performance of our proposed method under different parameter settings, we vary the parameters including: learning rate in range of $\{1e-4, 1e-3, 1e-2,1e-1\}$, number of epochs in the range of $\{100, 300, 500, 1000\}$, node representation dimension $d_z\in \{16, 32, 64, 128\}$, and historical embedding dimension $d_m\in \{16, 32, 64, 128\}$. From the results shown in Fig.~\ref{fig:parameter}, we observe that our method is generally not sensitive to parameter setting, but proper choices of parameters still benefit the performance.

\subsection{Case Study on Real-world Hospital Data}

\begin{table}[]
    \centering
 \caption{Estimated treatment effect of roommate number on MRSA infection in different populations of patients.}
 \def\arraystretch{1.1}%
    \begin{tabular}{cccc}
    \toprule
      \textbf{Population}   &  \textbf{T=0} & \textbf{T=1} & \textbf{T=2}\\
    \midrule
      \textbf{All}   & $0$ & $0.025 \pm 0.002$  & $0.082 \pm 0.004$  \\
      \hline
      \textbf{General Surgery} & $0$ & $0.016 \pm 0.002$  & $0.058 \pm 0.003$  \\
      \textbf{Intensive Care} & $0$ & $0.033 \pm 0.003$  & $0.119 \pm 0.005$  \\
      \textbf{Gerontology} & $0$ & $0.024 \pm 0.002$  & $0.082 \pm 0.004$  \\
    \bottomrule
    \end{tabular}
    \label{tab:MRSA_roommate}
\end{table}

\begin{table}[]
    \centering
\caption{Estimated treatment effect of hospital unit type on MRSA infection.}
\def\arraystretch{1.1}%
    \begin{tabular}{cccc}
    \toprule
      \textbf{Hospital Unit Type}   &  \textbf{Estimated ATE}\\
    \midrule
      \textbf{General Surgery}  & 0 (baseline)\\
      \textbf{Intensive Care} & $0.135 \pm 0.002$ \\
      \textbf{Gerontology} & $0.138 \pm 0.004$ \\
      \textbf{Transitional Care} & $-0.042\pm 0.006$ \\
      \textbf{Internal Medicine} & $0.000 \pm 0.002$\\
      \textbf{Cardiology} & $0.072 \pm 0.004  $ \\
        \textbf{Orthopedic Surgery} & $0.000 \pm 0.001$ \\
        \textbf{Gastroenterology} & $0.000 \pm 0.001$& \\
        \textbf{Hematology and Oncology} & $-0.083\pm 0.005$\\
    \bottomrule
    \end{tabular}
    \label{tab:MRSA_roomtype}
\end{table}
Methicillin-resistant Staphylococcus aureus (MRSA) is a difficult-to-treat pathogen (owing to multi-drug resistance) that is known to spread efficiently within hospitals via contact. One important avenue of hospitalized patient-to-patient MRSA transmission is thought to be through contamination of hospital room surfaces and equipment \cite{muto2003shea}. In addition, patients may be more or less susceptible to acquiring MRSA given individual factors \cite{shenoy2014natural}, and MRSA transmission rates may vary according to particular hospital wards (or hospital units) \cite{ohst2014network}.

The MRSA dataset contains observational data including patient covariates, room-sharing information, and MRSA test record from a real-world hospital. We construct a room-sharing network, in which an edge connects two patients (nodes) if and only if they have appeared in at least one same room simultaneously. We use our method to investigate the following causal questions: (1) How does the number of in-room contacts causally influence the MRSA infection risk? (2) How do other treatments, such as the type of hospital unit (e.g. Cardiology, Internal Medicine, etc.) causally influence the MRSA infection risk? 
As the ground-truth causal model is unknown, it is infeasible to evaluate our method on this dataset with the aforementioned metrics. Instead, we show some case studies and verify our key findings with domain knowledge.

For the first question, we map the number of in-room contacts into three levels of treatment. Here, treatments $0,1,2$ represent the roommate number from low to high. We take $T=0$ as the control group, and calculate the treatment effect for $T=1$ and $T=2$ by comparing the estimated potential outcomes of them with the case of $T=0$, respectively.
Table \ref{tab:MRSA_roommate} shows the estimated averaged treatment effect (ATE) of roommate number on MRSA infection over \textit{all} the patients, and also shows the estimated conditional averaged treatment effect (CATE) conditioned on each subpopulation of patients in a specific group of rooms. From the results, we observe that: 1) In general, the increase in roommate number could result in an increase in MRSA infection risk. This observation holds in the whole population and different subpopulations. As MRSA is contagious through physical contact, this observation is consistent with domain knowledge. 2) The CATE of roommate number on MRSA infection is the strongest in Intensive Care and Gerontology. In Intensive Care, it is frequent for patients to share devices such as ventilators, which leads to a more severe risk of infection when the number of in-room contacts increases. Besides, most patients in Gerontology rooms are older adults with comorbidities associated with MRSA susceptibility (i.e., age >79, prior nursing home residence, antibiotic exposure, dementia, stroke, or diabetes), which brings a higher risk for acquiring MRSA from the environment with more physical contact \cite{shorr2013risk}. 

For the second question, we take the hospital unit type as treatment, and show the estimated ATE of each hospital unit type on MRSA infection in Table \ref{tab:MRSA_roomtype}. Here, we take General Surgery as the baseline treatment (control group). From Table \ref{tab:MRSA_roomtype}, we observe that staying in Intensive Care and Gerontology rooms increases the MRSA infection risk most obviously. The reason might lie in the properties of these units (equipment sharing in the intensive care units, and more MRSA carriers in Gerontology). We also observe a relatively low treatment effect among beds in Transitional Care and Hematology/Oncology units. Most of these rooms are private (as opposed to other semi-private or 2-patient shared rooms), and may lead to less infection risk.
\vspace{-1mm}
\section{Related Work}
In this section, we introduce some representative studies related to this work, including causal inference on graph data and instrumental variable analysis.

\noindent\textbf{Causal inference on graph data.} Causal inference on graph data has recently attracted arising attention \cite{zheleva2021causal,guo2020learning,ma2021deconfounding,ogburn2020causal,wei2022causal}. Under this broad area, the topics which are most related to this work include: 
1) \textit{Entangled treatment}: 
a few initial explorations \cite{toulis2018propensity,toulis2021estimating} have been made for entangled treatment. These works discuss the challenges of entangled treatment modeling, and extend the traditional propensity score method for this problem. In our work, we do not limit the method to be propensity score-based, and consider a more general setting of entangled treatment with unknown treatment function, hidden confounders, and dynamic data. 
2) \textit{Network deconfounding}: A line of works \cite{guo2020learning,ma2021deconfounding} leverage the graph structure among units to capture the hidden confounders. Netdeconf \cite{guo2020learning} develops a GCN-based framework to learn the representations of hidden confounders, and adjusts for the confounders on top of the learned representations. DNDC \cite{ma2021deconfounding} further proposes to learn time-varying confounder representations from observational dynamic graphs. Although 
we also allow the existence of hidden confounders, our work differs from their application scenarios, as we focus on the setting in which the graph structure is an IV rather than a proxy for confounders. 
3) \textit{Network interference}: Traditional causal effect estimation studies are based on the Stable Unit Treatment Value (SUTVA) assumption \cite{rubin1980randomization,rubin1986statistics} that the treatment of each unit does not causally affect the outcome of other units (i.e., interference does not exist). However, interference often exists between connected units in graph data \cite{bhattacharya2020causal,zigler2021bipartite,aronow2017estimating}. There have been many works \cite{ma2021causal,aronow2017estimating,ma2022learning,yuan2021causal,hudgens2008toward,tchetgen2012causal} addressing the problem of causal inference under interference. Our work differs from them as we do not assume the existence of interference in graphs. Instead, we focus on the case when the graph influences the treatment assignment.

\noindent\textbf{Instrumental variable.} Hidden confounders can bring biases in causal effect estimation. Different from most causal inference methods which assume that all the confounders are observed, instrumental variable (IV) based methods provide an alternative approach to identifying causal effects even with the existence of hidden confounders. One of the most well-known lines of IV studies is two-stage methods  \cite{angrist2009mostly,hartford2017deep,darolles2011nonparametric,newey2003instrumental}. The two-stage least squares method (2SLS) \cite{angrist2009mostly} is the most representative work in this line, which first fits a linear model to predict treatment with features and IVs, and then fits another linear model to predict the outcome with the features and the predicted treatment. 2SLS is based on two strong assumptions: homogeneity (treatment effect is the same for different units) and linearity (the linear models are correctly specified). There have been many follow-up works to relax these assumptions. DeepIV \cite{hartford2017deep} is a neural network-based two-stage method that allows nonlinearity and heterogeneity. Another line of IV studies is based on the generalized method of moments (GMM) \cite{hansen1996finite,lewis2018adversarial}. Among them, DeepGMM \cite{bennett2019deep} leverages the moment conditions to identify the counterfactual generation function and estimate causal effects. But most of the existing IV studies focus on instrument variables in simple structures, such as scalars and vectors.

\section{Conclusion}
In this paper, {\red motivated from the task of investigating the impact of close contact on MRSA infection in a room-sharing network}, we studied the problem of causal effect estimation under entangled treatment. We discussed the related challenges and applications of this problem. To address this problem, we proposed a novel method \mymodel, which leverages the graph structure to better model the treatment assignments, and mitigates the confounding biases by using the graph structure as an instrumental variable. Considering the fact that the observational data is often time-varying in the real world, we further generalize the problem to a dynamic setting. Extensive experiments on synthetic, semi-synthetic, and real-world graph data validate the effectiveness of the proposed method. Especially, the validation of our method on real-world data provides insights for its future applications in real-world clinical studies. In the future, interesting directions of entangled treatment modeling on graphs include incorporating different levels of graph information (e.g., local-level and global-level) in treatment modeling, and considering entanglements in different types of graph data such as heterogeneous graphs and knowledge graphs. 

\section*{Acknowledgements}
This work is supported by the National Science Foundation under grants (IIS-1955797, IIS-2006844, IIS-2144209, IIS-2223769, CNS-2154962, BCS-2228534, CCF-1918656), the Commonwealth Cyber Initiative awards (VV-1Q23-007 and HV-2Q23-003), the JP Morgan Chase Faculty Research Award, the Cisco Faculty Research Award, the Jefferson Lab subcontract 23-D0163, the UVA 3 Cavaliers seed grant, the 4-VA collaborative research grant, the National Institutes of Health (NIH) grant 2R01GM109718-07, and CDC MIND cooperative agreement U01CK000589. One of the authors, Gregory Madden, is an iTHRIV Scholar. The iTHRIV Scholars Program is partly supported by the National Center for Advancing Translational Sciences of the National Institutes of Health under Award Numbers UL1TR003015 and KL2TR003016. This paper was prepared for informational purposes by the Artificial Intelligence Research group of JPMorgan Chase \& Co. and its affiliates (``JP Morgan''), and is not a product of the Research Department of JP Morgan. JP Morgan makes no representation and warranty whatsoever and disclaims all liability, for the completeness, accuracy or reliability of the information contained herein. This document is not intended as investment research or investment advice, or a recommendation, offer or solicitation for the purchase or sale of any security, financial instrument, financial product or service, or to be used in any way for evaluating the merits of participating in any transaction, and shall not constitute a solicitation under any jurisdiction or to any person, if such solicitation under such jurisdiction or to such person would be unlawful.

\clearpage
\bibliographystyle{ACM-Reference-Format}
\balance
\bibliography{ref}

\clearpage
\appendix
\section{Analysis}
In this section, we provide a more detailed analysis of the proposed method. 
Again, the outcome generation function defined in Eq. (\ref{eq:counterfactual}) is:
\begin{equation}
    Y^p = \mathcal{Y}(T^p, X^p, M^p) + g(U^p).
\end{equation}
Inspired by \cite{hartford2017deep}, a counterfactual prediction function is defined as:
\begin{equation}
    \mathcal{H}(T^p, X^p, M^p)  = \mathcal{Y}(T^p, X^p, M^p) + \mathbb{E}[g(U^p)|X^p, M^p].
\end{equation}
Here, $\mathcal{H}(T^p, X^p, M^p)$ is what we aim to estimate. As the hidden confounders $U^p$ cannot be observed, it is difficult for classical methods to directly fit this function from observational data. Fortunately, based on the assumptions mentioned in Section \ref{sec:assumption}, we have:
\begin{equation}
\begin{split}
    \mathbb{E}[Y^p|X^p, M^p, A^p] &= \mathbb{E}[\mathcal{Y}(T^p, X^p, M^p) + g(U^p)|X^p, M^p, A^p]\\
    &=\mathbb{E}[\mathcal{Y}(T^p, X^p, M^p)| X^p, M^p, A^p] \\ &\quad + \mathbb{E}[g(U^p)| X^p, M^p, A^p] \\
    &=\mathbb{E}[\mathcal{Y}(T^p, X^p, M^p)| X^p, M^p, A^p] \\ &\quad + \mathbb{E}[g(U^p)| X^p, M^p] \\
    &= \int \mathcal{Y}(T^p, X^p, M^p) d \mathcal{F}(T^p|X^p, M^p, A^p) \\ &\quad+
    \int\mathbb{E}[g(U^p)| X^p, M^p] d \mathcal{F}(T^p|X^p, M^p, A^p) \\ 
    &= \int \mathcal{H}(T^p, X^p, M^p) d \mathcal{F}(T^p|X^p, M^p, A^p),
\end{split}
\end{equation}
where $\mathcal{F}(T^p|X^p, M^p, A^p)$ is the conditional distribution of treatment. 
Here,  $\mathcal{H}$ can be estimated with an inverse problem based on observable functions $\mathbb{E}[Y^p|X^p, M^p, A^p]$ and $\mathcal{F}(T^p|X^p, M^p, A^p)$. In our two-stage IV analysis, the first stage can model $\mathcal{F}(T^p|X^p, M^p, A^p)$, and the second stage can model $\mathcal{H}(T^p, X^p, M^p)$.


\section{Details of Experiments}
In this section, we introduce more details of the experimental setup for the reproducibility of the experimental results. 

\subsection{Baseline Settings}
Here are more details for the settings of each baseline:
\begin{itemize}
    \item \textbf{S-Learner}: We use linear regression as the estimator in S-Learner.
    \item \textbf{Causal forest}: We set the number of trees as 100, the minimum number of samples required to be at a leaf node as 10, and the maximum depth of the tree as 10.
    \item \textbf{Counterfactual regression}: The number of epochs is set as $500$, the learning rate is $0.001$, the batch size is $4000$, the representation dimension is $25$. We choose Wasserstein-1 distance \cite{CFR} for representation balancing.
    \item \textbf{NetDeconf}: We set the number of epochs as 500, the learning rate as $0.005$, the representation dimension as $32$, and the representation balancing weight as $0.5$.
    \item \textbf{DNDC}: We set the number of epochs as 800, the learning rate as $0.001$, and the representation dimension as $32$.
    \item \textbf{DeepIV}:  We use the default parameter setting in the EconML package.
\end{itemize}

\subsection{Experiment Settings}
All the experiments in this work are conducted in the following environment:
\begin{itemize}
    \item Ubuntu 18.04
    \item Python 3.6
    \item Scikit-learn	1.0.1
    \item Scipy	1.6.2
    \item Pytorch 1.10.1
    \item Pytorch-geometric 1.7.0
    \item Networkx	2.5.1
    \item Numpy	1.19.2
    \item Cuda 10.1
\end{itemize}

\subsection{Dataset Details}
\noindent\textbf{Transaction.} This dataset is collected from the anti-money laundering (AML) financial system \cite{assefa2020generating,borrajo2020simulating} which provides  transaction records between users over time. At each timestamp, we construct a transaction network to represent the transactions occurring inside this timestamp. In the transaction network, each user is represented by a node, and a transaction is an edge between users. We use the user profiles such as location as their covariates. 

\noindent\textbf{Social.} This dataset contains a real-world social network of people at different timestamps based on tracking from smart devices \cite{de2013anatomy}. Each node represents a user, and each edge represents a friendship between two users. 

\subsection{Simulation Details}
By default, in our potential outcome simulation, we generate the elements in $\Theta_y$ from a random Gaussian distribution $\mathcal{N}(0, 0.1^2)$, and  $\Theta_0$ from $\mathcal{N}(0, 0.05^2)$. In the dynamic setting, we generate the weights $W_*^r$ with random sampling from $\mathcal{N}(1-(r/R), (1/R)^2)$. In this way, when $r$ is larger, the weights become smaller. This is consistent with the general observation that earlier information has weaker influence on the future data.

\end{document}